\documentclass[10pt, a4paper]{article}

\usepackage[]{lrec-coling2024} 
\usepackage{xurl}
\usepackage{enumitem}

\title{A Legal Framework for Natural Language Processing Model Training in Portugal}

\name{Rúben Almeida, Evelin Amorim} 

\address{INESC TEC, FCUP-Universidade do Porto \\
         ruben.f.almeida@inesctec.pt, evelin.f.amorim@inesctec.pt\\}

\abstract{
Recent advances in deep learning have promoted the advent of many computational systems capable of performing intelligent actions that, until then, were restricted to the human intellect. In the particular case of human languages, these advances allowed the introduction of applications like ChatGPT that are capable of generating coherent text without being explicitly programmed to do so. Instead, these models use large volumes of textual data to learn meaningful representations of human languages. Associated with these advances, concerns about copyright and data privacy infringements caused by these applications have emerged. Despite these concerns, the pace at which new natural language
processing applications continued to be developed largely outperformed the introduction of new regulations. Today, communication barriers between legal experts and computer scientists motivate many unintentional legal infringements during the development of such applications. In this paper, a multidisciplinary team intends to bridge this communication gap and promote more compliant Portuguese NLP research by presenting a series of everyday NLP use cases, while highlighting the Portuguese legislation that may arise during its development.
\newline \Keywords{Portuguese NLP, Legal NLP, PLN Português} }

\begin{document}

\maketitleabstract

\section{Introduction}
\label{sec:introduction}

In recent years, deep-learning-based methods have permitted great advances in Computer Science~(CS) fields previously known for their computational complexity~\cite{bishop_bishop_2023}. One of those fields was natural language processing~(NLP); the CS field focused on machine understanding of human languages and the generation of coherent human text~\cite{deng_liu_2018}.

Long before the introduction of state-of-the-art~(SOTA) generative large language models~(LLMs) like Chat-GPT~\cite{openai_2022}, NLP relied on simpler approaches based on rules/word counts~\cite{jurafsky_martin_2014} to deliver human-interpretable models capable of addressing simple NLP tasks. The evolution towards deep learning permitted NLP to quit the controlled atmosphere of university labs and embrace mainstream societal adoption at the expense of costly training processes and vast volumes of textual data. 

The capabilities revealed by SOTA NLP models were accompanied by ethical and legal concerns among prominent NLP researchers\footnote{\url{https://twitter.com/ylecun/status/1733481002234679685}}, big-tech CEOs\footnote{\url{https://www.reuters.com/technology/musk-experts-urge-pause-training-ai-systems-that-can-outperform-gpt-4-2023-03-29/}}, politicians\footnote{\url{https://www.ft.com/content/9339d104-7b0c-42b8-9316-72226dd4e4c0}}, and economists\footnote{\url{https://www.weforum.org/agenda/2017/03/taxing-robots-wont-work-says-yanis-varoufakis/}} who appealed for regulation and higher ethical standards during the development of NLP solutions as a way of restricting the usage of such capabilities to induce harm in society.

However, the pace at which new LLMs are currently being developed largely surpasses the pace at which new regulations are introduced. This phenomenon, together with the traditional communication barriers between the formality of legal expertise and the dynamic world of CS, opens the door to legal vacuums, promoting undesirable copyright\footnote{\url{https://www.nytimes.com/2023/12/27/business/media/new-york-times-open-ai-microsoft-lawsuit.html}} and privacy infringements\footnote{\url{https://www.forbes.com/sites/emmawoollacott/2023/09/01/openai-hit-with-new-lawsuit-over-chatgpt-training-data/?sh=2f3d1b856d84}}. 

In this paper, we intend to bridge this gap by listing the legal concerns that may arise during everyday NLP challenges. We focus on the Portuguese legal system to present a study targeting both computer scientists and legal experts, with the ultimate goal of promoting awareness about the topic in one of the most peripherally and underdeveloped\footnote{\url{https://en.wikipedia.org/wiki/List_of_countries_by_Human_Development_Index}} countries of the European Union~(EU). Despite our focus on Portugal, we believe researchers in other EU countries may find the concepts introduced in this paper useful, due to the European nature of many of the legislation listed in this paper.

The paper is structured as follows: Section~\ref{sec:related_work} describes the existing literature about the topic. The lack of research, focused exclusively on the Portuguese case, forced us to broaden our scope towards other EU countries. Section~\ref{sec:portuguese_nlp} provides a quick overview of the current SOTA for Portuguese NLP, with a special emphasis on explaining the importance of Brazilian NLP research for Portuguese NLP. In Section~\ref{sec:portuguese_legal}, we briefly introduce the Portuguese legal system, performing a high-level description of the legislation that may engage with the development of NLP solutions. Section~\ref{sec:licenses} lists common licensing agreements used by NLP researchers to publish their works. Finally, in Section~\ref{sec:use_cases}, we introduce some use cases representing everyday challenges faced by NLP researchers highlighting the legal concerns associated with them.

\section{Related Work}
\label{sec:related_work}

The novelty of this subject translates into a lack of literature about the topic. The absence would be even worse if we focused exclusively on the Portuguese legal landscape. Therefore, in the following section, we broaden our scope outside Portugal towards other EU countries.

\subsection{Portuguese}

The existing literature concerning the legal implications of artificial intelligence~(AI) systems in Portugal is constrained to regional conferences and journals in the Portuguese language. We highlight the pioneering works of~\cite{guimaraes2021direito} that compiled summarized versions of master theses developed by law students about different CS topics, including Big Data~\cite{silva_costa_2021}.

Recently,~\cite{barbosa_2023} introduced the analysis concerning data privacy during LLM development. Despite focusing on the Portuguese legal framework, the writer relies heavily on EU regulations to provide a brief overview of the recent developments surrounding the topic. In particular, it is demonstrated that the General Data Protection Regulation~(GDPR) alone is incapable of regulating LLMs like ChatGPT. The author mentions the importance of the AI Act to establish a legal framework founded on the concept of \textbf{risk} to deal with the inaccuracies of these models. 

Focusing on copyright issues, we highlight the works of~\cite{nobre_2012}, which provide an extensive listing of what is protected and what is excluded from copyright protection.

Lastly, it is worth mentioning the interesting considerations of~\cite{pereira_2020} regarding copyright eligibility for AI outputs. In Portugal, the copyrights for AI results are free, unlike in other countries like the United Kingdom, where the research team who made them inherits the rights\footnote{\url{https://pec.ac.uk/blog_entries/copyright-protection-in-ai-generated-works/}}.

\subsection{European}

At the European level, we highlight the works of~\cite{eckart2018legal} and ~\cite{kelli2020impact} focusing on copyright. Both works discuss whether an LLM trained in copyrighted protected data inherits the same license as its training data. Both authors concluded there is a lack of clarity on the topic, but it all depends on how much copyright data is used. If a model can reproduce data that is protected by copyright, it is called a derivation of the dataset and has the same copyright as the dataset used.

Nevertheless, most of the European literature about the topic is produced by enterprises. We highlight the efforts of ML6\footnote{\url{https://www.ml6.eu/blogpost/navigating-ethical-considerations-developing-and-deploying-large-language-models-llms-responsibly}}, KPMG\footnote{\url{https://kpmg.com/ie/en/home/insights/2024/01/eu-artificial-intelligence-act-art-int.html}} or EY\footnote{\url{https://www.ey.com/en_ch/forensic-integrity-services/the-eu-ai-act-what-it-means-for-your-business}} to promote the debate about the topic. These blog posts often have anonymous authors who are not checked by others. This makes it hard to know if they are accurate and up-to-date.

\section{Portuguese NLP: Quick Overview}
\label{sec:portuguese_nlp}

Portuguese is the official language of 260 million people spread across five continents. In the context of NLP, Portuguese is considered a mid-resourced language~\cite{joshi2021state}. Meaning that ``they have a large amount of unlabeled data...and are only challenged by lesser amount of
labeled data''~\cite{joshi2021state}. Despite this classification, the vast majority of these resources are Brazilian Portuguese, produced by Brazilian research teams.

Similar to other mid-resource languages, the shorter investment produces a reduced number of Portuguese LLMs. Until recently, BERTimbau~\cite{souza2020bertimbau}, a Brazilian Portuguese version of BERT, was the only Portuguese LLM. Progressively, more complex architectures have emerged: Albertina PT~\cite{rodrigues2023advancing} or the generative models Sabiá~\cite{pires2023sabia}, Gervásio~\cite{santos2024advancing}, and GlórIA~\cite{lopes2024gloria}. 

\subsection{Leveraging Brazilian Portuguese Resources}

Despite the major phonological, morphological, lexical, syntactical and semantic differences between the numerous Portuguese varieties around the globe~\cite{scherre2016main}, prompt-engineering\footnote{\url{https://en.wikipedia.org/wiki/Prompt_engineering}} and fine-tuning\footnote{\url{https://en.wikipedia.org/wiki/Fine-tuning_(deep_learning)}} of Brazilian LLMs help European Portuguese researchers achieve SOTA results~\cite{almeida2024indexing}.
\section{Portuguese Legal System}
\label{sec:portuguese_legal}

Portugal, as an EU member, must ensure its legal system is in harmony with EU law. In this section, we list the legislation that engages with NLP development, while describing how EU legislation impacts the Portuguese legal system.

\subsection{Scientific Exceptions}

Many of the legislation listed in the following subsections establish exceptions to scientific work. For that reason, we believe it is paramount to clarify what is considered \textbf{scientific work} by the Portuguese law.

The~\cite{EUdirective2019790} directive introduces a comprehensive definition of scientific work founded on the concept of profit: ``research organisation means a university, including its libraries, a research institute or any other entity, the primary goal of which is to conduct scientific research or to carry out educational activities involving also the conduct of scientific research:
(a)~on a not-for-profit basis or by reinvesting all the profits in its scientific research; or
(b)~pursuant to a public interest mission recognised by a Member State"~\cite{EUdirective2019790}.

\subsection{National Legislation}

Portuguese legislation is divided into two major sets of laws: the civil code and the penal code. The penal code lists those actions that constitute a crime and require the existence of \textbf{willful misconduct} to be taken into consideration. While in the civil code, the penalties restrained themselves to some kind of compensation, the penal code establishes tougher penalties like prison time. Committing a crime is a severe action that usually implies not only a penal condemnation but also a parallel civilian compensation to repair any harm provoked in society.

Regarding NLP, most of the legislation that applies is restricted to the civil code; however, in recent years, with the mass adoption of CS in society, many penal considerations have arisen focusing on copyright infringement, privacy violations, or cybercrime.

Finally, before enumerating the Portuguese legislation directly relevant to NLP, it is essential to clarify the distinctions between public, semi-public, and private crimes. Public crimes warrant the intervention of public prosecutors to safeguard the interests of the victim. Semi-public crimes trigger public prosecution only upon the filing of a complaint with the police. In cases of private crimes, typically of lesser severity, public prosecution does not intervene, and a complaint must also be lodged with the police. In instances where the penal code does not explicitly specify the classification of the crime, it defaults to being considered public\footnote{\url{https://www.ministeriopublico.pt/perguntas-frequentes/crime}}.

\begin{description}[style=unboxed, leftmargin=0px]
    
    \item[Sensitive Data Protection:] Article 35 of the Portuguese constitution~\cite{PortugueseConstitutionArt35} strictly prohibits the digital processing of sensitive ethnic, political, sexual, or religious data without explicit consent and for scientifically motivated purposes that may positively impact society. Consequently, NLP researchers are advised to refrain from operating on such sensitive data.
    
    \item[Right to Expression:] Article 79 of the Portuguese civil code~\cite{CodigoCivilArt79} affirms the right of individuals to freely express themselves without fear of their words being recorded or utilized by a third party. This legislation outlines exceptions for scientific research. Without these provisions, the use of tweets to train NLP models would be deemed illegal. 
    
\end{description}

\subsection{European Legislation}

The great investment the EU has been making to regulate AI has had a great effect in peripheral countries like Portugal, helping reduce inequalities in access to technology between richer and poorer countries within the EU.

Before enumerating EU regulations that engage directly with the development of NLP solutions, it is important to clarify some terminology regarding EU law:

\begin{description}[leftmargin=0px]
    
     \item[Regulation:] Regulations come into force automatically upon approval by the European Parliament and European Council. Requiring a 65\% majority for acceptance, regulations are always accompanied by a transition period to allow EU citizens to adapt to the new legal framework.

    \item[Directive:] Directives represent EU laws that member states must adopt within a specified transition period, integrating them into their own legal systems. Despite penalties for countries failing to timely adopt directives, Portugal has earned a reputation for delays in transposing these directives. A pertinent example is the 2019/790 directive, where the transposition was delayed by over two years\footnote{\url{https://www.cuatrecasas.com/pt/portugal/art/transposicao-da-diretiva-do-mercado-unico-digital-1}}. 

\end{description}

Below, we briefly introduce the most relevant EU legislation that engages with NLP development:

\begin{description}[style=unboxed, leftmargin=0px]
    \item[General Data Protection \textbf{(Regulation)}:] Enacted in 2016, this EU regulation~\cite{EUregulation2016679} sets the foundational standards for online data protection. Emphasizing transparency and consent, it mandates that all corporate operations involving personal data must be pre-informed and explicitly consented to by users. The regulation applies specifically to private data - information that, if disclosed, can identify the individual to whom it pertains. It also regulates cross-border data transfers, stipulating that EU data protection laws accompany the data wherever it goes. Outside the EU, in countries lacking an \textbf{adequacy decision}\footnote{\url{https://www.edpb.europa.eu/sme-data-protection-guide/international-data-transfers_en}}, such as Brazil, additional measures may be necessary to ensure equivalent levels of protection as those mandated within the EU. This regulation was integrated into Portuguese legislation in 2021, complemented by the introduction of the \textbf{Portuguese Charter on Human Rights in the Digital Age}~\cite{cartaPTDHED}. Concerning NLP, explicit consent is required for data collection and usage in training NLP models, with researchers urged to minimize the use of private data whenever possible. While scientific research exceptions exist, the legislature advises their avoidance, favoring maximal consent.

    \item[Copyright in the Digital Single Market \textbf{(Directive)}:] The 2019/790 EU directive~\cite{EUdirective2019790} on copyrights introduced a \textbf{right to~[text]~mine}, framing the legal context for NLP development. It broadened exceptions, permitting scientists to use copyrighted data for NLP model training if not pursued for profit. Transposed into Portuguese legislation in 2023~\cite{DecretoLei472023}, this directive provides a legal framework for NLP endeavors.
    
    \item[Database Sui Generis Protection \textbf{(Directive)}:] Directive 96/97EC~\cite{directive969EC} established copyright protection for databases, granting 15 years of copyright protection to those who compile independent works into structured units. This regulation was revisited during the approval of the 2019/790 directive, with exceptions allowing copyright infringements for scientific endeavors also applying to databases.
    
    \item[AI Act \textbf{(Regulation)}:] Recently approved, this regulation~\cite{COM2021206} aims to be the world's first comprehensive AI legal framework, positioning the EU at the forefront of AI legislation. It sets rules for AI systems, including NLP, based on their societal risk levels. Additionally, it mandates transparency mechanisms during the training of LLMs exceeding $10^{25}$ FLOPS. Given that most NLP research does not involve sensitive data or yield automated decisions significantly impacting Europeans' daily lives, it is categorized as \textbf{minimal-risk}, exempting researchers from additional considerations. The AI Act is in the final stages of approval and is expected to be fully implemented within two years.

\end{description}

\section{NLP Licensing System}
\label{sec:licenses}

In Table~\ref{tab:copyrights}, we list the licenses of many NLP resources.

\begin{table}[!h]
    \begin{tabular}{lcc}
        \hline
        \textbf{Model Name}    & \textbf{Year} & \textbf{License}                   \\
        \hline
        BERT & 2018 & Apache 2.0 \\
        GPT-2 & 2019 & MIT \\
        Bloom & 2022 & Custom License \\
        Falcon & 2023 & Apache 2.0 \\
        Llama 2 & 2023 & Custom License \\
        Mistral & 2023 & Apache 2.0 \\
        Phi-2 & 2023 & MIT \\
        Gemma & 2024 & Custom License \\
        \hline
        \hline
        BERTimbau     & 2020 & MIT                       \\
        Albertina PT  & 2023 & MIT                       \\
        Sabiá         & 2023 & Llama 1 License           \\
        Gervásio      & 2024 & MIT                       \\
        \textbf{Glória}  & \textbf{2024} & \textbf{ClueWeb22 License} \\
        \hline
        \hline
        Wikitext      & 2016 & CC-BY-SA 3.0             \\
        CNN-Dailymail & 2017 & Apache 2.0                \\
        Flores        & 2019 & CC-BY-4.0                 \\
        OSCAR         & 2023 & CC0-1.0                   \\
        \hline
    \end{tabular}
    \caption{Licenses for different SOTA NLP resources. The first entries cover non-Portuguese LLMs, while the second set focuses on Portuguese architecture. The last set of resources are commonly used NLP datasets.}
    \label{tab:copyrights}
\end{table}

The results reveal that many LLMs adopt an Apache 2.0 or MIT license. The datasets identified tend to adopt Common Crawl licenses. It is worth mentioning the case of the Portuguese LLM Glória~\cite{lopes2024gloria}, where the usage of the clueweb22 dataset~\cite{overwijk2022clueweb22} as part of the training corpus required Glória's authors to adopt this license as well.

The information provided in this section is complemented by the extensive analysis made by the choose a license platform\footnote{\url{https://choosealicense.com/appendix/}}.
\section{Use Cases}
\label{sec:use_cases}

In this section, we cover three use cases that represent everyday NLP tasks. In the following subsections, we assume the perspective of a Portuguese NLP operating under EU law. The green color used in the flowcharts represents the yes/permitted case; in contrast, the red color represents the no/not permitted case.

\subsection{Load Brazilian Portuguese Dataset From HuggingFace}
\label{use_case:1}

\begin{figure}[!ht]
    \centering
    \includegraphics[width=\linewidth]{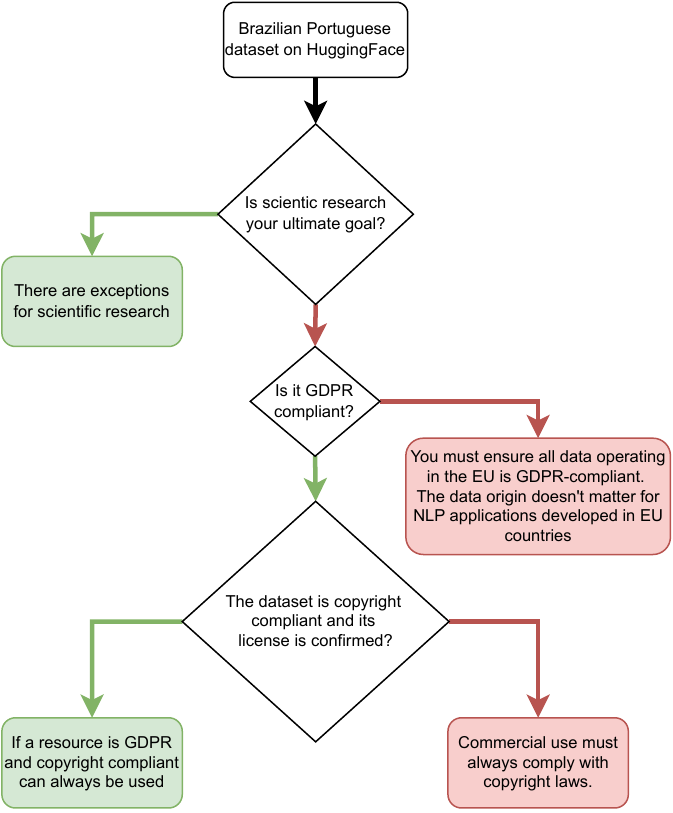}
    \caption{Flowchart summarizing the legal questions associated with the loading of a Non-EU dataset}
    \label{fig:use_case_1}
\end{figure}

This use case describes a standard practice in Portuguese NLP, where Brazilian datasets are used to train new NLP models. In Figure~\ref{fig:use_case_1} we provide a flowchart summarizing the legal questions that may arise during the process.

It is worth mentioning that the geographical origin of the dataset has no impact on the overall legal assessment of the resource.

\subsection{Crawl Portuguese Websites to Produce a Large NLP Corpus}
\label{use_case:2}

\begin{figure}[!ht]
    \centering
    \includegraphics[width=\linewidth]{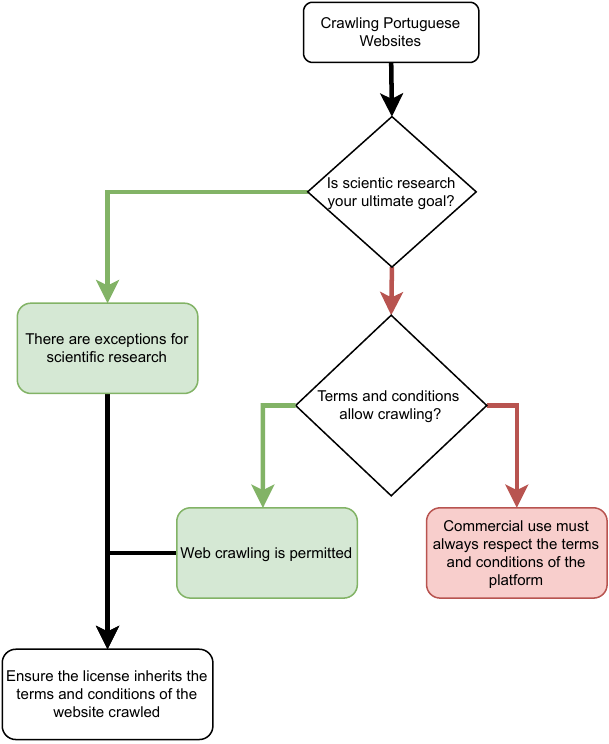}
    \caption{Flowchart highlighting the legal considerations web crawling may arise.}
    \label{fig:use_case_2}
\end{figure}

The usage of web crawling techniques is paramount for LLM training. SOTA LLMs require a vast amount of data, whose scale is only comparable to the amount of information existing on the web. In Figure~\ref{fig:use_case_2}, we describe the legal considerations researchers should pay attention to while crawling websites.

\subsection{Use Tweets to Produce Political Profiles: The Facebook-Cambridge Analytica case}
\label{use_case:3}

\begin{figure}[!ht]
    \centering
    \includegraphics[width=\linewidth]{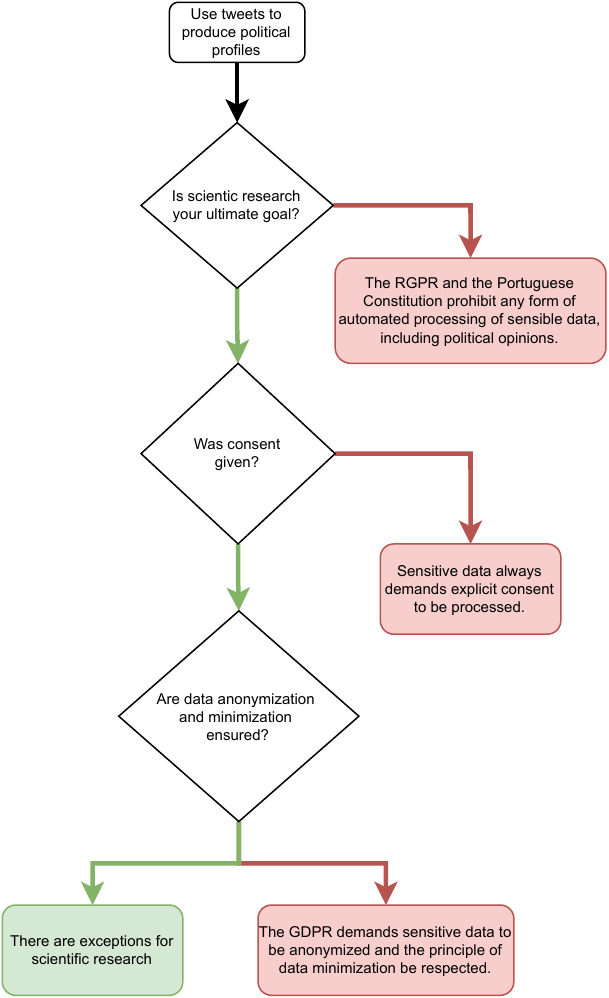}
    \caption{Flowchart concerning the legal issues of processing sensitive data.}
    \label{fig:use_case_3}
\end{figure}

This use case draws inspiration from the Facebook–Cambridge Analytica data scandal\footnote{\url{https://shorturl.at/uxK47}}. It aims to encapsulate the legal considerations that may emerge during the development of NLP models, particularly in scenarios involving sensitive data such as political profiling. Figure~\ref{fig:use_case_3} provides a summarized overview of these legal aspects.

\section{Conclusion \& Future Work}

In this study, we have presented the inaugural legal framework for NLP development in Portugal. The limited awareness of this subject in one of the EU's less affluent nations exposes vulnerabilities to legal vacuums and potential infringements of copyright and data privacy. We have tried to introduce key NLP concepts in a way that is accessible to both computer scientists and legal experts. The three use cases outlined serve to summarize the principal insights gained from this research. Utilizing flowcharts to illustrate these scenarios aims to accelerate the procedure of obtaining data and enhance overall comprehension of the research.

Our future work will focus on expanding the scope of use cases, matching to the same structured approach, while exploring additional topics. Specifically, we aim to incorporate insights from the recently introduced regulation on AI, the European AI Act, to further enrich the legal framework for NLP development.

\section{Acknowledgements}

This work is financed by National Funds through the Portuguese funding agency, FCT - Fundação para a Ciência e a Tecnologia, within project LA/P/0063/2020.
DOI 10.54499/LA/P/0063/2020 | https://doi.org/10.54499/LA/P/0063/2020
The author also would like to aknowledge the project StorySense, with reference 2022.09312.PTDC (DOI 10.54499/2022.09312.PTDC).

\nocite{*}
\section{Bibliographical References}\label{sec:reference}

\bibliographystyle{lrec-coling2024-natbib}
\bibliography{references}

\end{document}